\documentclass[letterpaper, 10 pt, conference]{ieeeconf}
\IEEEoverridecommandlockouts
\usepackage[pdftex]{graphics}
\usepackage{epsfig}
\usepackage{times}
\usepackage{amsmath}
\usepackage{amssymb}
\usepackage{bm}
\usepackage[colorlinks,linkcolor=red,anchorcolor=blue,citecolor=green]{hyperref}
\usepackage{subfigure}
\usepackage[T1]{fontenc}
\usepackage[linesnumbered,ruled,vlined]{algorithm2e}
\usepackage[noend]{algpseudocode}
\usepackage{booktabs}
\usepackage{array}
\usepackage{color}
\usepackage{multirow}

\usepackage{float}
\usepackage{xcolor}
\usepackage{url}
\usepackage{cite}
\usepackage[export]{adjustbox}
\usepackage{balance}

\begin{document}

\title{\LARGE \bf
PTT: Point-Track-Transformer Module for 3D Single Object Tracking in Point Clouds
}

\author{Jiayao Shan$^\dagger$, Sifan Zhou$^\dagger$, Zheng Fang*, Yubo Cui
	\thanks{$^\dagger$Authors with equal contribution.}
	\thanks{The authors are with the Faculty of Robot Science and Engineering, Northeastern University, Shenyang, China; Corresponding author: Zheng Fang, e-mail: fangzheng@mail.neu.edu.cn}
	}	

\maketitle

\begin{abstract}
3D single object tracking is a key issue for robotics. In this paper, we propose a transformer module called Point-Track-Transformer (PTT) for point cloud-based 3D single object tracking. PTT module contains three blocks for feature embedding, position encoding, and self-attention feature computation. Feature embedding aims to place features closer in the embedding space if they have similar semantic information. Position encoding is used to encode coordinates of point clouds into high dimension distinguishable features. Self-attention generates refined attention features by computing attention weights. Besides, we embed the PTT module into the open-source state-of-the-art method P2B to construct PTT-Net. Experiments on the KITTI dataset reveal that our PTT-Net surpasses the state-of-the-art by a noticeable margin (\textcolor{red}{$\sim$10\%}). Additionally, PTT-Net could achieve real-time performance ($\sim$40FPS) on NVIDIA 1080Ti GPU. Our code is open-sourced for the robotics community at \url{https://github.com/shanjiayao/PTT}.
\end{abstract}

\section{INTRODUCTION}
\label{sec:introduction}

\textit{3D single object tracking (SOT)} has a wide range of applications in robotics and autonomous driving \cite{kartobject,machida2012human}. However, most existing 3D SOT methods are equipped with RGB-D cameras \cite{context,RGBDtracker}, which inherit the characteristics of 2D images and depend heavily on RGB-D information, thus trackers may fail in visually degraded or illumination changing environments.

In addition to RGB-D sensors, 3D LIDAR sensors are also widely used in object tracking tasks \cite{FaF,Complexer-YOLO} because they are less sensitive to illumination changes and could directly capture geometric information more accurately. However, using only point clouds for 3D SOT has its challenges. \textit{First}, point cloud is sparse and disordered \cite{PointNet}, which requires the network to be permutation-invariant. \textit{Second}, 3D object tracking needs to estimate higher space dimension (e.g. $x,y,z,w,h,l,ry$) than 2D visual tracking, which takes more computational complexity. \textit{Third}, compared with rigid object tracking (e.g. car), it's more challenging to track non-rigid objects (e.g. pedestrian) since it is too hard to extract stable features.

\begin{figure}[t]
	\centering
	\setlength{\abovecaptionskip}{-10pt}
	\includegraphics[width=\linewidth]{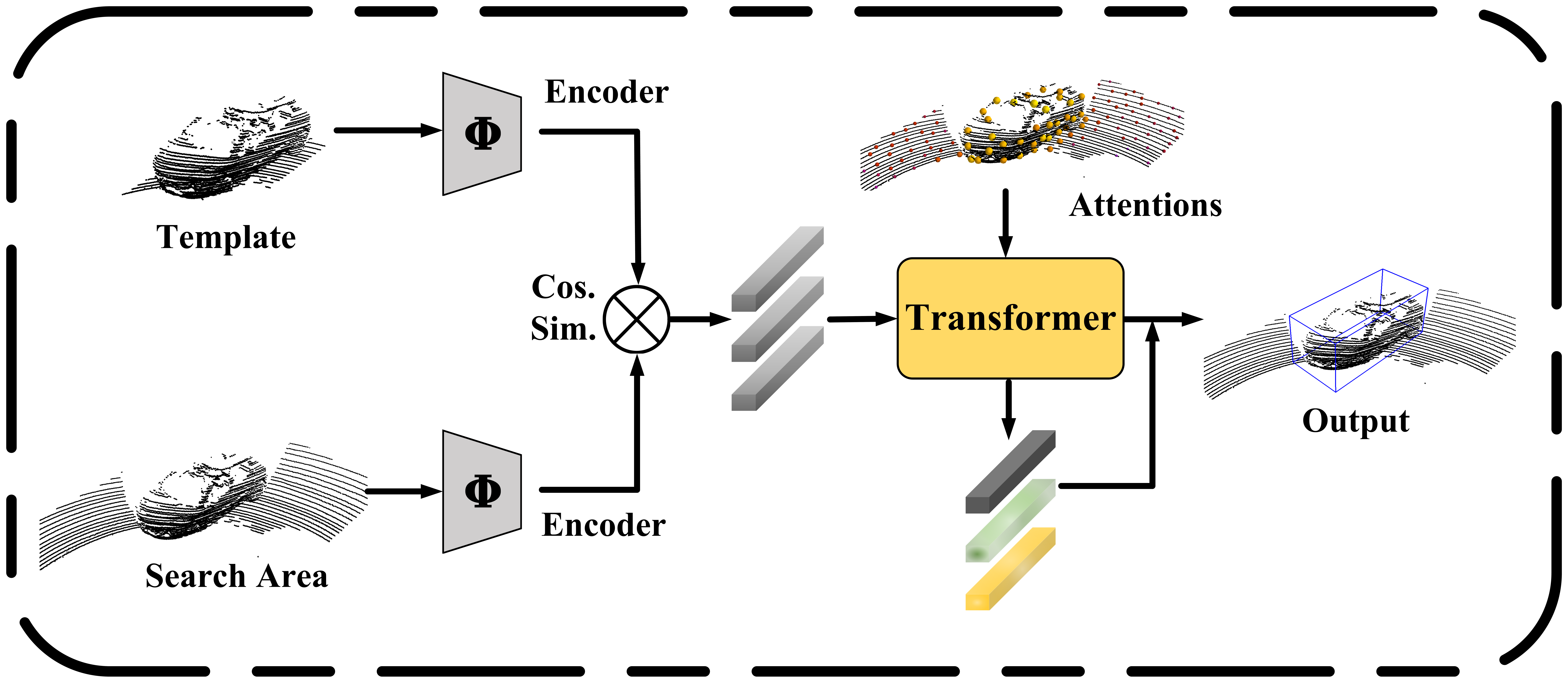}
	\caption{\textbf{Exemplified illustration to show how PTT module works}. Compared with the existing 3D single object tracking method, our PTT module works after calculating the similarity features, and weighs the features based on their importance to improve tracking performance.}    
	\label{fig:abstract}
\end{figure}

Currently, most point cloud based 3D SOT methods follow \textit{2D visual object tracking (VOT)} approaches (e.g. Siamese networks \cite{SiamFC,SiameseRPN}) to track 3D target. The Siamese network formulates VOT task as learning a similarity function between the template branch and the search branch. SC3D \cite{SC3D} is the pioneer point cloud based 3D Siamese Tracker based on shape completion network. However, it could not be trained end-to-end and run in real-time. Besides, Qi et al. \cite{P2B} also proposed a point-to-box (P2B) network to estimate target bounding box from the raw point cloud. However, this approach does not cope well with sparse scenarios. Recently, Fang et al. \cite{3DSiamRPN} jointed Siamese network and point cloud based Region Proposal Network (RPN) \cite{PointRCNN} to tackle 3D SOT task. Nonetheless, the performance of their method is unsatisfying. It is worth noting that points located in different geometric positions often have different importance in representing targets. However, these aforementioned methods do not weigh point cloud features based on this characteristic. Besides, the point cloud features extracted from the template and the search area contain less potential object information due to the sparsity of point clouds. Therefore, how to pay attention to the important clues is the key to improving the performance of the 3D object tracker.

Recently, transformer has revolutionized natural language processing and image analysis \cite{vaswani2017,hu2019local,ramachandran2019stand}. Self-attention operator, which is the core of transformer networks, is intrinsically a set operator: positional information is provided as attributes of elements that are processed as a set \cite{vaswani2017,zhao2020san}. Therefore, transformer is suitable for point cloud due to its positional attributes.

In this paper, we explore the application of the transformer network for 3D SOT task, and propose a transformer module named PTT (Point-Track-Transformer). To pay more attention to the important feature of the object, we use the transformer's powerful self-attention and position encoding mechanism to weigh the point cloud features. To evaluate the effect of our PTT module, we embed our PTT module into the open-source state-of-the-art method P2B \cite{P2B} to build a new network called PTT-Net. Finally, the experimental results of our PTT-Net on KITTI \cite{kitti} dataset demonstrate the superiority of our method ($\sim$10\%’s improvement on both Success and Precision). Besides, PTT-Net could run at 40FPS.

Overall, there are three main contributions as follows:

$\bullet$ \textbf{PTT module}: a Point-Track-Transformer (PTT) module for 3D single object tracking using only point clouds, weighing point cloud features to focus on deeper-level object clues during tracking.

$\bullet$ \textbf{PTT-Net}: a 3D single object tracking network named PTT-Net embedded with PTT modules which can be end-to-end trained. To the best of our knowledge, this is the first work to apply transformer to 3D object tracking task based on point cloud.

$\bullet$ \textbf{Open-source}: Experiments on KITTI tracking dataset \cite{kitti} show our method outperforms the state-of-the-art methods with remarkable margins. Besides, we open source our method to the research community.

\section{RELATED WORK}
\label{sec:related-works}

This section will briefly discuss the related work in 3D single object tracking, transformer and self-attention mechanism.

\subsection{3D SOT Using Point Cloud}

Giancola et al. \cite{SC3D} proposed the first pioneer point cloud based 3D single object tracker which utilized the Kalman Filter to generate massive target proposals. They introduced the shape completion module to enrich the feature learning on points. However, their method has a poor generalization ability and could not run in real-time. Zarzar et al. \cite{Zarzar2019} leveraged 2D Siamese network which converted raw point clouds into Bird-Eye-View (BEV) representation to generate 3D proposals. This method may lose fine-grain geometry details which are important for tracking 
tiny objects. Cui et al. \cite{Cui2019} also adopted a 3D Siamese tracker only using point cloud. However, they could not estimate the orientation and size information of the target. Fang et al. \cite{3DSiamRPN} jointed 3D Siamese network and 3D RPN network to track targets, but their performance is limited by the one-stage RPN network. Besides, Zou et al. \cite{FSiamese} integrated 2D image and 3D point cloud information for 3D SOT. However, this method relies more on 2D trackers and the performance is not satisfactory when using previous results to initialize tracker. Qi et al. \cite{P2B} proposed P2B which used deep hough voting to obtain the potential centers (votes) and estimated target center based on those votes. However, it ignores the fact that points in different positions have different contributions to tracking. Furthermore, its random sampling mechanism loses the location distribution information of the raw point cloud. Based on these shortcomings, we propose a PTT module to weigh different point features and use farthest point sampling instead of random sampling to obtain more raw point cloud information.

\subsection{Transformer and Self-attention}

Recently, there have been many wonderful works based on Transformer \cite{vaswani2017,hu2019local}. Hu et al. \cite{hu2019local} and Ramachandran et al. \cite{ramachandran2019stand} applied scalar dot product self-attention in local pixel neighbors. Zhao et al. \cite{zhao2020san} utilized vector self-attention operations to image tasks.

Inspired by these works, Zhao et al. \cite{Jia22020} used a Point Transformer layer by applying vector self-attention operations, which had a great performance improvement in point cloud classification and segmentation tasks. Nico et al. \cite{Engel2020} proposed SortNet as a part of Point Transformer, and achieved competitive performance on point cloud classification and partial segmentation tasks. Meanwhile, Guo et al. \cite{Guo2020} also introduced Point Cloud Transformer (PCT), which performed well on shape classification, part segmentation, and normal estimation tasks. Obviously, transformer has unique advantages for point cloud feature learning. We hence aim to extend the transformer paradigm to our 3D SOT task with more attention to features.

\begin{figure*}[ht]
	\centering
	\setlength{\abovecaptionskip}{-30pt}
    \setlength{\belowcaptionskip}{-10pt}
	\includegraphics[width=\linewidth]{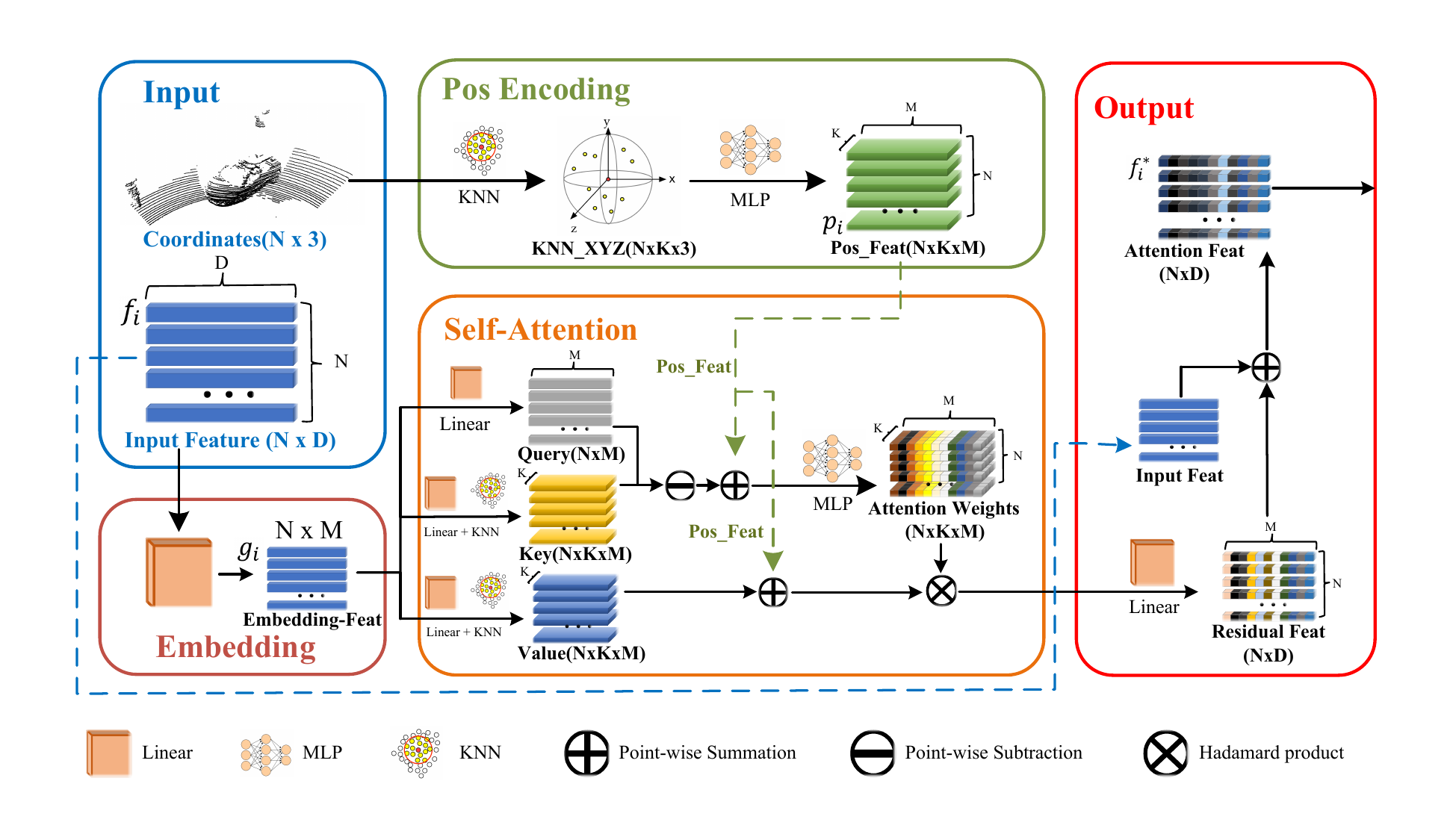}
	\caption{\textbf{PTT module architecture.} It consists of three blocks: feature embedding, position encoding, and self-attention. The whole inputs are the coordinates and their corresponding features. Feature embedding module maps input features into embedding space. In position encoding module, the k-nearest neighbor algorithm is used to obtain local position information, then the encoded position features will be learned by a Multi Layer Perceptron (MLP). The self-attention module learns refined attention features for input features based on local context. The output features of PTT module are the sum of input and residual features.}
	\label{fig:transformer}
	\vspace{-10pt}
\end{figure*}

\section{METHODOLOGY}
\label{sec:method}

Here, we focus on the application of transformer network in 3D SOT. Given an input of M points with XYZ coordinates, a backbone network modified on the basis of PointNet++ \cite{PointNet++} is used to extract the point cloud and learn deep features. It outputs a subset of the input containing N interest points (seeds) ${S=\left\{s_i\right\}^{N}_{i=1}}$. ${s_i=(c_i,f_i)}$ is composed by a vector ${c_i}$ of 3D coordinate and a D-dimensional descriptor ${f_i}$ of the local object geometry. Our goal of using transformer is to perform an attention weighting operation on the feature space of $f_i$, and output refined features ${f_i}^*$ with the same dimension.

\subsection{Transformer}
The architecture of transformer can be divided into three parts: input feature embedding, position encoding, and self-attention. Self-attention is the core module, which mainly focuses on the differences of input features and generates refined attention features based on global or local context. Given the input feature ${G=\left\{g_i\right\}^{N}_{i=1}}$ after feature embedding, the general formula of self-attention is:
\begin{equation}
    \begin{aligned}
        &Q,K,V = \alpha(G), \beta(G), \gamma(G)\\
        &\mathcal{A} = \rho(\sigma(Q, K) + P) \odot (V)
    \end{aligned}
\label{eq:general-trans}
\end{equation}
where $\alpha$, $\beta$ and $\gamma$ are point-wise feature transformations (e.g. linear layers or MLPs). $Q$, $K$, and $V$ are the $query$, $key$ and $value$ matrices, respectively. $\sigma$ is the relation function between $Q$ and $K$. $P$ is the position encoding feature. $\rho$ is a normalization function (e.g. \textit{Softmax}). $\odot$ means Hadamard product.  $\mathcal{A}$ is the attention feature produced by self-attention layer.

\subsection{PTT Module}

We modify the transformer module proposed in Point Transformer \cite{Jia22020} to weigh the point cloud features. In \cite{Jia22020}, the point transformer layer is proposed to process the raw point cloud for classification and segmentation tasks. Here, we set an importance-based transformer module to focus on the differences among the input point cloud features for 3D SOT task. 

Therefore, we explore how to embed transformer modules into 3D SOT task and propose our Point-Track-Transformer (PTT) module. PTT module processes features by utilizing shape and geometry information. Given a point set ${S=\left\{s_i\right\}^{N}_{i=1} }$, $s_i=(c_i,f_i)$, $c_i \in \mathbb{R}^{3}$ and $f_i \in \mathbb{R}^{D}$. ${c_i}$ and ${f_i}$ represent 3D coordinates and descriptor of point $s_i$.

Feature embedding module maps input features into embedding space $\mathbb{R}^{M}$: $f_i \rightarrow g_i$, $g_i \in \mathbb{R}^{M}$. Position encoding module extracts higher-level M-dimensional features $p_i$ from input coordinates $c_i$: $c_i \rightarrow p_i$, $p_i \in \mathbb{R}^{K \times M}$. Finally, the self-attention module calculates attention weights and attention features $f_i^*$,$f_i^* \in \mathbb{R}^{D}$ by taking embedding features and position features as inputs. To avoid the vanishing gradient problem in training stage, we also adopt the residual architecture in \cite{Jia22020}, and take the sum of the attention features and input features as output features.

\subsubsection{Feature Embedding} 

The original feature embedding module in Natural Language Processing (NLP) is to map each word in the input sequence to a high-dimensional vector. In this work, we use the linear layer to complete the feature embedding operation, and map the input point cloud feature dimension from $D$ To $M$: $\mathbb{R}^{D} \rightarrow \mathbb{R}^{M}$, which can place the feature closer in the embedding space if the semantics are more similar and make the network have a stronger fitting ability.

\subsubsection{Position Encoding} 

Position encoding module plays a crucial role in transformer, which allows operators to adapt to the local structure of the input data \cite{vaswani2017}. And 3D point coordinates themselves are the natural input for position encoding. Therefore, we utilize the coordinates directly as the input of the position encoding module. Besides, we use the relative coordinates to make the network better capture the spatial correlation between points and local geometric shape information. Since the feature $f_i$ is extracted by \cite{PointNet++} which can provide the local context information, we obtain the position encoding features ${P=\left\{p_i\right\}^{N}_{i=1}}$ with function $\eta$. For input point set $S$ including N points, the position encoding feature for each point is:
\begin{equation}\label{eq:pos-encode}
    {p_i}= \eta(c_i - c_j)
\end{equation}

where $c_i$ is the coordinate of the i-th point in $S$. $c_j$ is the j-th coordinate in local neighborhood region of $c_i$ by using K-nearest-neighbor (KNN). And $\eta$ is an MLP with two linear layers and one ReLU layer.

\begin{figure*}[t]
	\centering
	\setlength{\abovecaptionskip}{-17pt}
	\includegraphics[width=\linewidth]{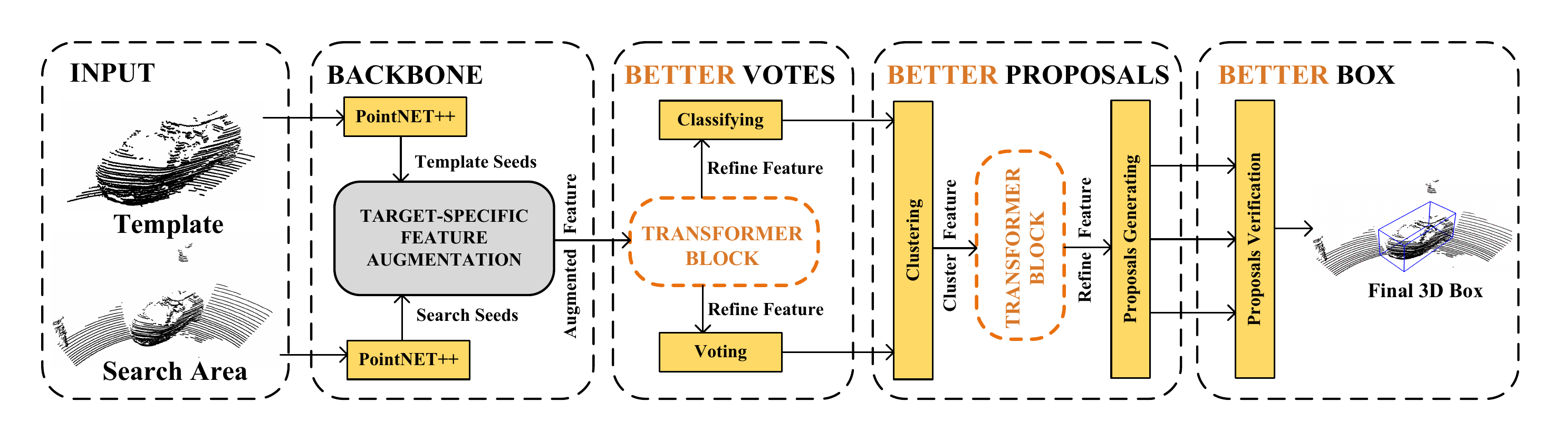}
	\caption{\textbf{The pipeline of PTT-Net.} In order to verify the effect of our PTT module, we embedded two PTT modules into seeds voting stage and proposals generation stage of \cite{P2B}. The first transformer block takes the augmented feature as input, and outputs the refine features for voting. The second transformer block refines the cluster features to help PTT-Net obtain better proposals.}
	\label{fig:main-pipeline}
	\vspace{-10pt}
\end{figure*}

\subsubsection{Self-Attention} 

As Fig.~\ref{fig:transformer} shows, self-attention module computes three vectors for each point: $Q$, $K$, $V$ through $\alpha$, $\beta$, $\gamma$, where $\alpha$, $\beta$, $\gamma$ are all linear layers. It is worth noting that K and V are aggregated from the features of the $k$ neighborhood points, which aim to encode more local context information. Here, $Q \in \mathbb{R}^{M}$, $K \in \mathbb{R}^{K \times M}$, and $V \in \mathbb{R}^{K \times M}$. 

The relation function $\sigma$ can be classified into two types: scalar \cite{vaswani2017} and vector \cite{zhao2020san}. And it has been proved in \cite{Jia22020} that vector attention is more suitable for point cloud than scalar attention since it supports adaptive modulation of individual feature channels, not just whole feature vectors. Thus we set $\sigma(Q, K)=Q - K$ to obtain point-wise attention weights. And an MLP layer $\gamma$ is used to introduce additional trainable transformations and match the output dimension. Then, we add the position encoding features $P$ to both the attention vector $\sigma$ and the transformed features $K$. Finally, the residual features recorded as $\mathcal{A}$ are defined as the weighted sum of the attention weights with all $V$ vectors. The formula is as follows:
\begin{equation}\label{eq:attention}
\mathcal{A} = \rho\big(\gamma(Q - {K} + {P})\big) \odot \big({V} + {P}\big)
\end{equation}

where $\rho$ is a normalization function (Softmax) and $\gamma$ is a non-linear mapping function (MLP) that includes two linear layers and one ReLU layer. $\mathcal{A}$ is attention features.

\subsection{PTT-Net}
\label{sec:ptt-net}

The ability of the transformer to learn self-attention weights inspires us to try it on 3D SOT task. We formulate the problem of focusing on the differences in features as self-attention weighting. In order to verify the effect of our method, we embed our PTT module into P2B \cite{P2B}. More specifically, the PTT modules are inserted in seeds voting stage and proposals generation stage of P2B.

In seeds voting stage, P2B generates votes based on the augmented features, which are from backbone in Fig.~\ref{fig:main-pipeline}. We notice that \cite{P2B} ignores the differences among different point cloud features in the search area, and gives no preference to the points in different locations when generating votes. However, it is important to focus on the points which contain more geometric information and suppress the background noise. Therefore, we apply PTT module to weigh the augmented features and obtain the weighted features focused on foreground points (in Fig.~\ref{fig:transformer_show}(a)(b)).

In proposals generation stage, P2B generates proposals based on local context features. However, their method ignores the global semantic features of targets, so that they could not distinguish similar objects (e.g. two pedestrians, in Fig.~\ref{fig:transformer_show}(c)(d)). Therefore, we use the PTT module to further weigh the target-wise context features obtained by the aggregation network in P2B for tracking deeper-level target clues. 

As shown in Fig.~\ref{fig:main-pipeline}, we embed our PTT module in the P2B \cite{P2B} to build PTT-Net. We add PTT module to the seeds voting and proposal generation stages, and weigh the augmented features and cluster features respectively. Experiments show that our PTT-Net outperforms \cite{P2B} with remarkable margins.

\subsubsection{Loss Function}

The PTT module is trained with the other subnetworks in \cite{P2B}. So we follow \cite{P2B} to design our loss function. The overall loss consists of two parts as follows:
\begin{equation}
\begin{aligned}
&{L}_{all} = {L}_{cv} + \lambda_{1}{L}_{cb}+\lambda_{2}{L}_{rv}+\lambda_{3}{L}_{rb}
\end{aligned}
\label{equ:loss}
\end{equation}
where $\lambda_{1}$, $\lambda_{2}$, $\lambda_{3}$ represent the weighting coefficient of each loss. Classification loss includes voting classification loss ${L}_{cv}$ and proposal box classification loss ${L}_{cb}$. The regression loss includes the voting loss ${L}_{rv}$ and the proposal box regression loss ${L}_{rb}$.

\section{EXPERIMENTS}
\label{sec:experiment}

We used KITTI tracking dataset \cite{kitti} as the benchmark, and set up more detailed experiments to show the superiority of our PTT-Net by comparing the different performance of the dominant methods \cite{SC3D,P2B,3DSiamRPN,FSiamese} on rigid (Car,Van) and non-rigid (Pedestrian,Cyclist) objects. The experiments show that PTT-Net outperforms the previous SOTA method with remarkable margins at 40fps. 

\subsection{Experimental protocols}
\subsubsection{Dataset}
We used the training set of KITTI which includes more than 20,000 manually labeled 3D objects using Velodyne HDL-64E 3D lidar (10HZ). For the splits of dataset, we follow \cite{SC3D,P2B,3DSiamRPN,FSiamese}, which divide 20 sequences into three parts 00-16, 17-18, 19-20, corresponding to training set, validation set, and test set respectively.

\subsubsection{Evaluation Metric}
Following previous work \cite{SC3D,P2B,3DSiamRPN,FSiamese}, we report Success and Precision metrics defined by One Pass Evaluation (OPE) \cite{OPE}, which represent overlap and error Area Under the Curve (AUC) respectively.

\subsubsection{Implementation Details}
We use the farthest point sampling (FPS) instead of random sampling (RS) in origin P2B \cite{P2B}. In training stage, we use the Adam optimizer and set the initial learning rate to 0.001 and decrease by 5 times after 12 epochs. The batch size is 48 and training epoch is 60. Besides, we extend the offset from (x, y, $\theta$) to (x, y, z, $\theta$) when generating more template samples during data augmentation in \cite{P2B}. In testing stage, we also add $Z$ axis offset to generate predicted box. Other parameters are consistent with settings of \cite{P2B}. 

\subsection{Quantitative Experiments}

\begin{table}[t!]
\vspace{4pt}
\renewcommand\arraystretch{1.3}
    \begin{center}
    \setlength{\belowdisplayskip}{-10pt}
    \caption{Performance comparison on the KITTI dataset for the car category. R and L denote RGB ans LiDAR respectively.}
    \label{tab:evaluation_car}
    \setlength{\tabcolsep}{5pt}{
	\begin{tabular}{ccccc}
		\toprule
		Module  & Modality & 3D Success & 3D Precision & FPS \\\hline
		AVOD-Tracking\cite{AVOD}  & R+L & 63.1 & 69.7 & -\\
		F-Siamese\cite{FSiamese}  & R+L & 37.1 & 50.6 & -\\\hline
		SC3D\cite{SC3D}  & L & 41.3 & 57.9 & 1.8\\
		P2B\cite{P2B}  & L & 56.2 & 72.8 & {\color{red} 45.5}\\
		3D-SiamRPN\cite{3DSiamRPN}  & L & {\color{blue} 58.2}$^2$ & {\color{blue} 76.2} & 20.8\\
		\textbf{PTT-Net(Ours)} & L & {\color{red} 67.8}$^1$ & {\color{red} 81.8} & {\color{blue} 40.0}\\
		\hline
    \end{tabular}}
    \end{center}
    $^1$ $^2$Red and blue mean the performance score is ranked first and second respectively.
\end{table}

\begin{table}[t!]
\renewcommand\arraystretch{1.3}
\begin{center}
\caption{Extensive comparisons with different categories. ``Ped" denotes ``Pedestrian". }
\label{tab:evaluation_KITTI}
\begin{tabular}{lcccccc}
\toprule
\multirow{2}{*}{}          & Category                                                & Car  & Ped & Van  & Cyclist & Mean  \\ 
                           & \begin{tabular}[c]{@{}c@{}}Frame  Number\end{tabular} & 6424 & 6088      & 1248 & 308     & 14068 \\ \hline
\multirow{5}{*}{\rotatebox{90}{Success}}   & SC3D\cite{SC3D}                                                    & 41.3 & 18.2      & 40.4 & \color{blue}{41.5}    & 31.2  \\
                           & P2B\cite{P2B}                                                     & 56.2 & 28.7      & 40.8 & 32.1    & 42.4  \\
                           & FSiamese\cite{FSiamese}                                               & 37.1 & 16.2      & -    & \color{red}{47.0}    & -     \\
                           & 3DSiamRPN\cite{3DSiamRPN}                                              & \color{blue}{58.2} & \color{blue}{35.2}      & \color{red}{45.6} & 36.1    & \color{blue}{46.6}  \\
                           & \textbf{PTT-Net(Ours)}                                           & \color{red}{67.8} & \color{red}{44.9}      & \color{blue}{43.6} & 37.2    & \color{red}{55.1}  \\ \hline
\multirow{5}{*}{\rotatebox{90}{Precision}} & SC3D\cite{SC3D}                                                    & 57.9 & 37.8      & 47.0 & \color{blue}{70.4}    & 48.5  \\
                           & P2B\cite{P2B}                                                     & 72.8 & 49.6      & 48.4 & 44.7    & 60.0  \\
                           & FSiamese\cite{FSiamese}                                               & 50.6 & 32.2      & -    & \color{red}{77.2}    & -     \\
                           & 3DSiamRPN\cite{3DSiamRPN}                                              & \color{blue}{76.2} & \color{blue}{56.2}      & \color{red}{52.8} & 49.0    & \color{blue}{64.9}  \\
                           & \textbf{PTT-Net(Ours)}                                           & \color{red}{81.8} & \color{red}{72.0}      & \color{blue}{52.5} & 47.3    & \color{red}{74.2}  \\ \hline
\end{tabular}
\end{center}
\vspace{-0.2cm}
\end{table}

To better evaluate our method, we designed two quantitative experiments on KITTI dataset. In the first experiment, we quantitatively evaluated our method for 3D car tracking. In the second experiment, we further compared PTT-Net with the previous methods on Pedestrian, Van, and Cyclist.

\subsubsection{Comparisons on car category}
We compared the performance of our PTT-Net with the existing methods on the KITTI dataset and reported results for 3D car tracking in Tab.~\ref{tab:evaluation_car}. In order to fit the requirement of real scenarios, we generate the search area centered on the previous result. The results show our PTT-Net has achieved SOTA performance in all evaluation metrics. Compared with the baseline algorithm P2B \cite{P2B}, our performance has been greatly improved by $\sim$11\%. Additionally, compared with \cite{AVOD} and \cite{FSiamese} which both use RGB+LIDAR fusion information, the Success/Precision results of PTT-Net outperform them 4.7\%/12.1\% and 30.7\%/31.2\% respectively.

\subsubsection{Comparisons on other categories}

We also compared with the dominant methods on Pedestrian, Van, and Cyclist (Tab.~\ref{tab:evaluation_KITTI}). The average performance of PTT-Net outperforms P2B \cite{P2B} $\sim$13\%. It is worth noting that the Success/Precision results of PTT-Net show an improvement (9.7\%/15.8\%) on non-rigid object (Pedestrian) tracking. This also proves that our PTT module can help the network understand and learn the important characteristics of the target better.

\subsection{Ablation Study}
\label{sec:ablation-study}

\begin{table}[]
    \vspace{4pt}
    \renewcommand\arraystretch{1.3}
    \begin{center}
    \setlength{\belowdisplayskip}{-10pt}
    \caption{Different ways for template generation. "GT" denotes "ground truth". "First \& Previous" denotes "The first ground truth and Previous result". }
    \begin{tabular}{lccccc}
    \toprule
    \multirow{2}{*}{}          & \multirow{2}{*}{Method} & \multirow{2}{*}{\begin{tabular}[c]{@{}c@{}}The First\\ GT\end{tabular}} & \multirow{2}{*}{\begin{tabular}[c]{@{}c@{}}Previous \\ Result\end{tabular}} & \multirow{2}{*}{\begin{tabular}[c]{@{}c@{}}First \& \\ Previous\end{tabular}} & \multirow{2}{*}{\begin{tabular}[c]{@{}c@{}}All \\ Previous\end{tabular}} \\
                               &                         &                                                                                   &                                                                             &                                                                               &                                                                          \\ \hline
    \multirow{4}{*}{\rotatebox{90}{Success}}   & SC3D\cite{SC3D}                    & 31.6                                                                              & 25.7                                                                        & 34.9                                                                          & 41.3                                                                     \\
                               & P2B\cite{P2B}                       & 46.7                                                                              & 53.1                                                                        & 56.2                                                                          & 51.4                                                                     \\
                               & 3DSiamRPN\cite{3DSiamRPN}                & 57.2                                                                              & -                                                                           & 58.2                                                                          & -                                                                        \\
                               & \textbf{PTT-Net(Ours)}             & \textbf{62.9}                                                                              & \textbf{64.9}                                                                        & \textbf{67.8}                                                                          & \textbf{59.8}                                                                     \\ \hline
    \multirow{4}{*}{\rotatebox{90}{Precision}} & SC3D\cite{SC3D}                      & 44.4                                                                              & 35.1                                                                        & 49.8                                                                          & 57.9                                                                     \\
                               & P2B\cite{P2B}                       & 59.7                                                                              & 68.9                                                                        & 72.8                                                                          & 66.8                                                                     \\
                               & 3DSiamRPN\cite{3DSiamRPN}                & 75.0                                                                              & -                                                                           & 76.2                                                                          & -                                                                        \\
                               & \textbf{PTT-Net(Ours)}           & \textbf{76.5}                                                                              & \textbf{77.5}                                                                        & \textbf{81.8}                                                                          & \textbf{74.5}                                                                     \\ \hline
    \end{tabular}
    \end{center}
    \vspace{-0.2cm}
    \label{tab:different_template}
\end{table}

\begin{table}[t!]
\renewcommand\arraystretch{1.2}
    \begin{center}
    	\centering
        \caption{Different embedded positions of PTT module.}
    	\setlength\tabcolsep{15.0pt}{
    	\begin{tabular}{ccc} 
    	    \toprule
            \textbf{~Ablation} & 3D Success & 3D Precision \\
            \midrule
            baseline\cite{P2B}  & 56.2 & 72.8 \\
            Only PTT in Vote & 62.1& 76.9 \\
            Only PTT in Prop & 65.7 & 78.9 \\
            PTT in all(PTT-Net) &\pmb{67.8} & \pmb{81.8} \\ \hline
        \end{tabular}}
    \end{center}
	\vspace{-0.2cm}
	\label{tab:Ablation-ptt-pos}
\end{table}

\begin{figure}[t]
	\centering
	\setlength{\abovecaptionskip}{-15pt}
	\includegraphics[width=\linewidth]{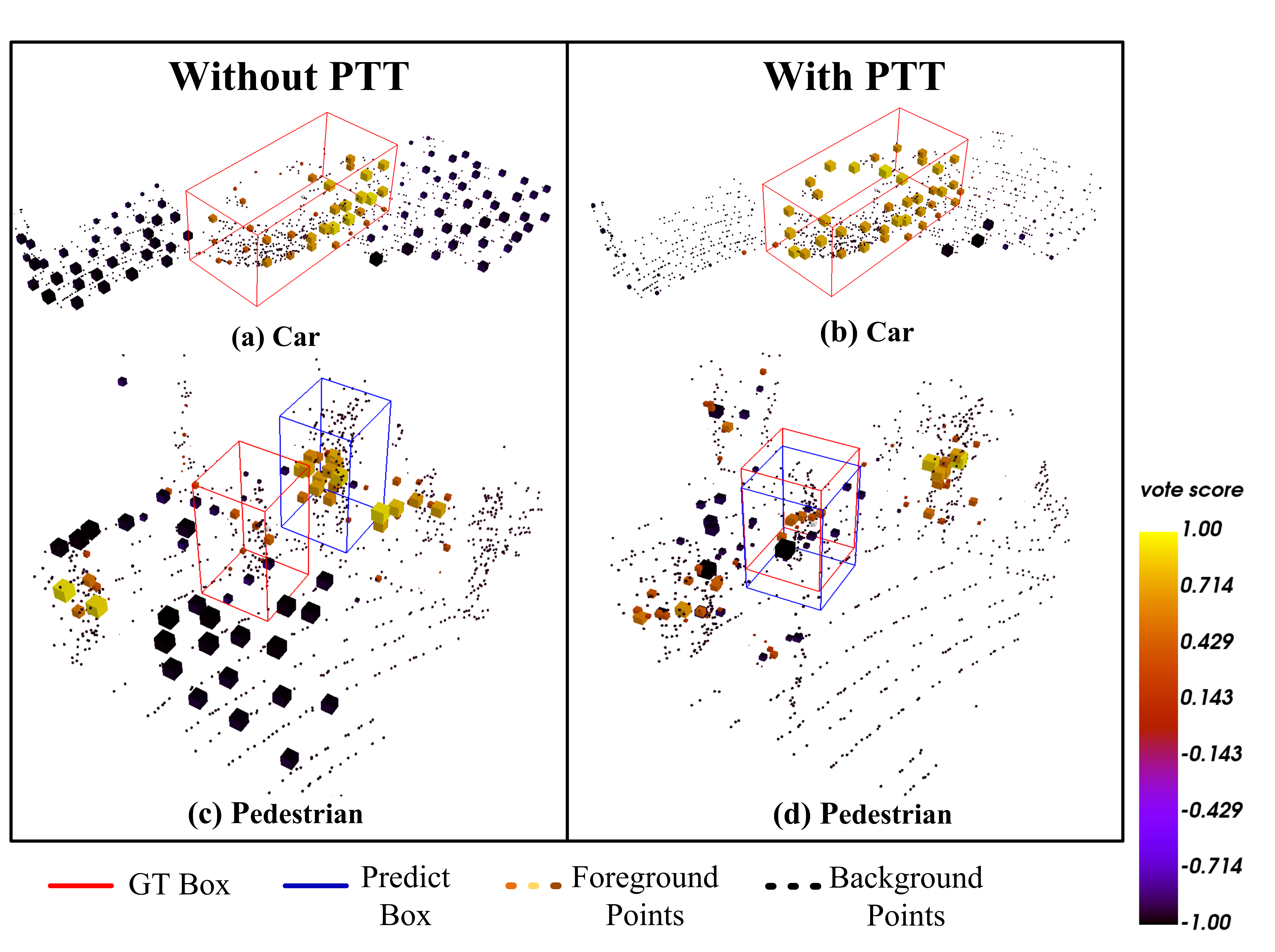}
    \caption{\textbf{Visualization of classification(a-b) and tracking(c-d) results with or without PTT module}. The point will be paid more attention if it has a higher score. Compared (a) with (b), PTT module pays more attention to the foreground points. Compared (c) with (d), PTT module could still track targets robustly in crowded scenes (with multiple pedestrians).}
	\label{fig:transformer_show}
\end{figure}

\begin{figure}[t]
	\centering
	\setlength{\abovecaptionskip}{-15pt}
	\includegraphics[width=\linewidth]{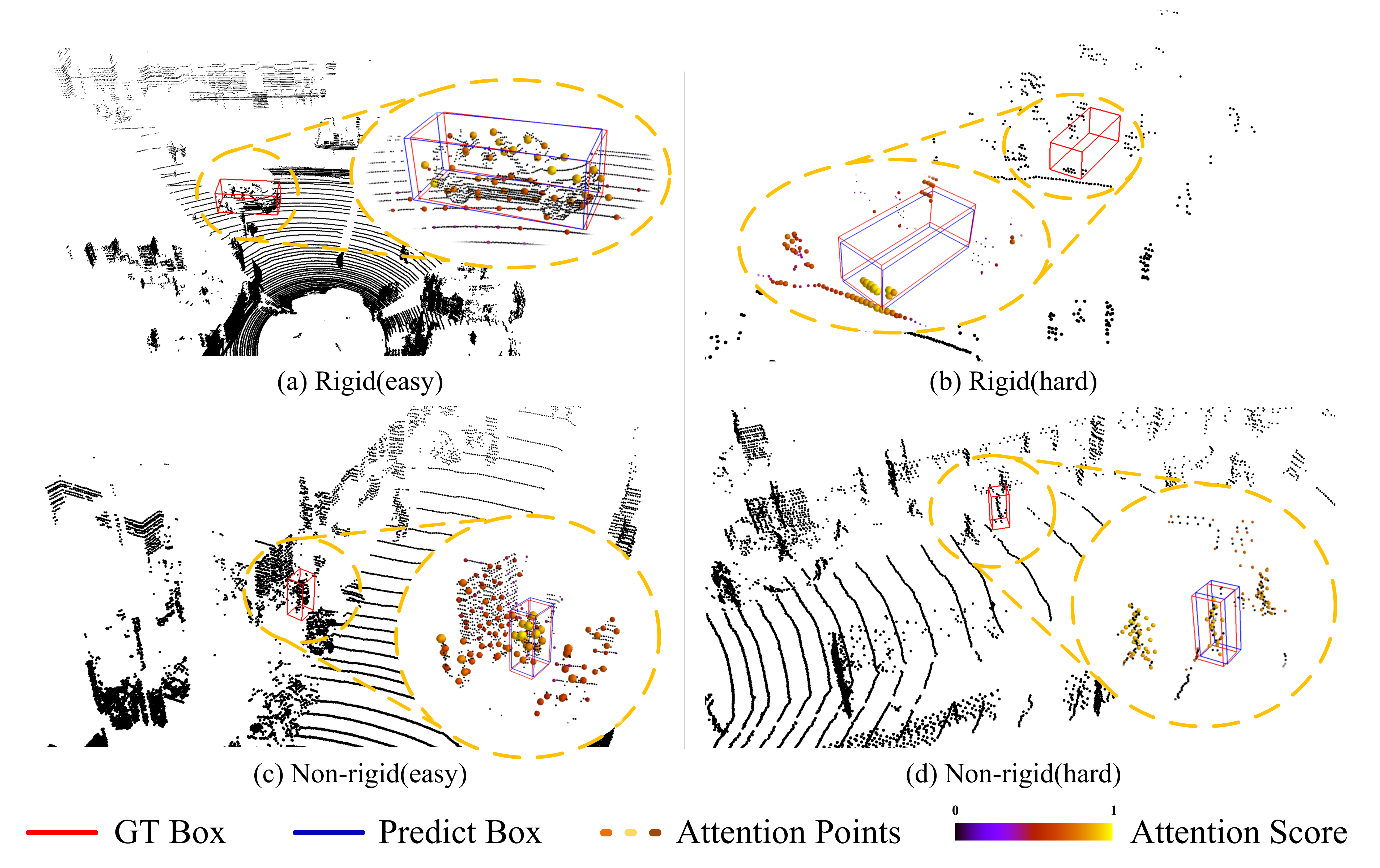}
	\caption{\textbf{Exemplified illustration to show the attention scores of rigid cases(a-b) and non-rigid cases(c-d).} And we also divide the easy and hard cases according to the number of foreground points. It is obvious that PTT module guides tracker in focusing on foreground points even if they are few or extremely similar to others.}
	\label{fig:attention-show}
\end{figure}

\subsubsection{Template Area Generation}
We explored four different settings of template point cloud generation, including the first ground truth, previous result, the fusion of the first ground truth and previous result, and all previous results. We reported results in Tab.~\ref{tab:different_template}. Obviously, our method achieves SOTA performances in all settings.

\subsubsection{Embedding location of PTT module}
To verify our design in Sec.~\ref{sec:ptt-net} of positions where PTT modules are embedded, we tried different schemes (Tab.~\ref{tab:Ablation-ptt-pos}). The results show that embedding PTT module in both two stages of \cite{P2B} can obtain the best improvement. Besides, as shown in Fig.~\ref{fig:transformer_show}, compared with (a) and (b), PTT-Net has better point cloud classification results which focus on foreground points. Comparing (c) with (d), PTT-Net could still track target pedestrian robustly when more proposal centers are generated from another pedestrian. This result effectively shows that transformer can learn more target-wise information.

\begin{figure}[t]
	\centering
	\setlength{\abovecaptionskip}{-7pt}
	\includegraphics[width=0.9\linewidth]{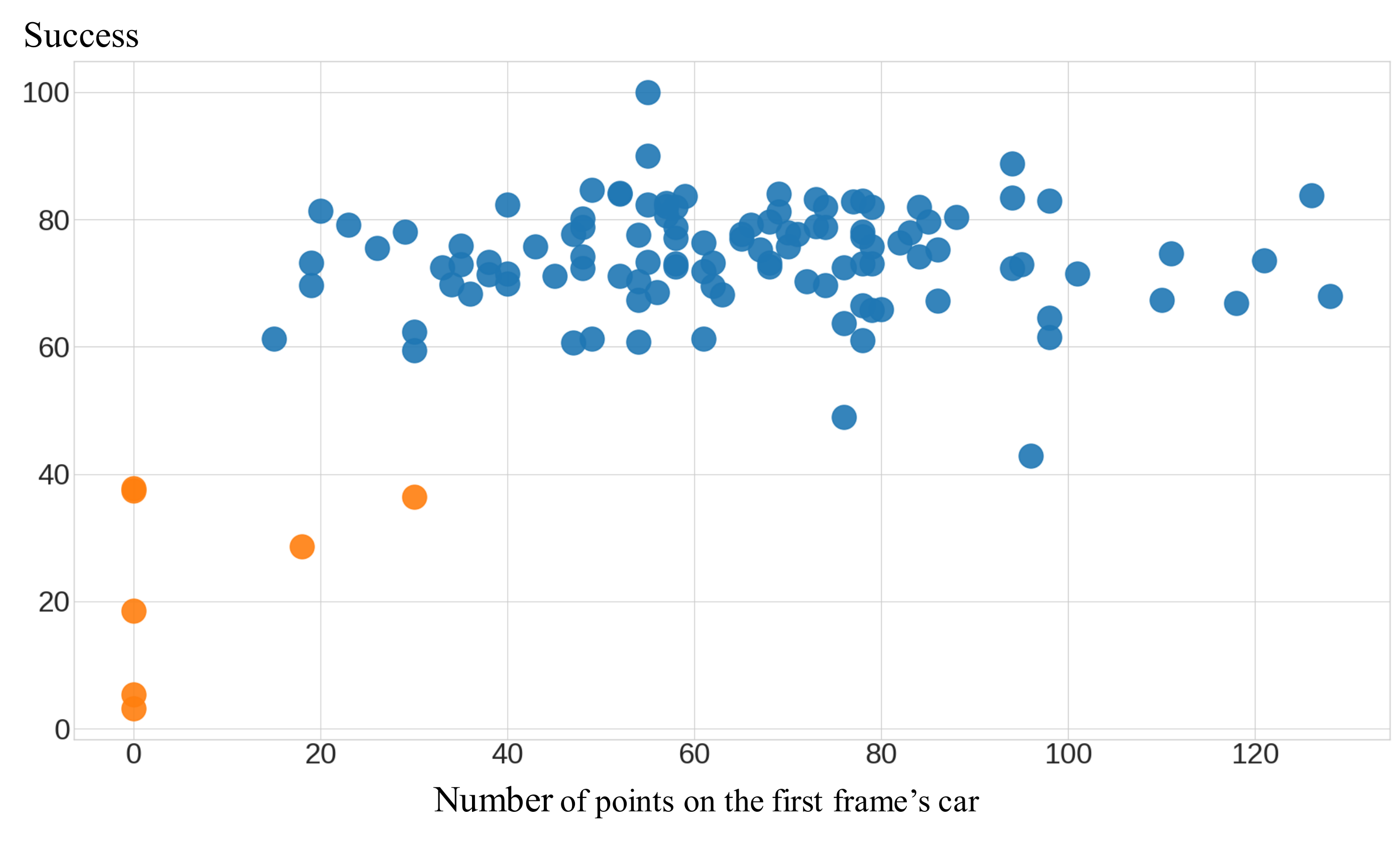}
	\caption{\textbf{The number of points on the first frame's car and 3D Success}. There are 120 points in this scatter figure corresponding to 120 tracklets in testing sequences (19-20). Orange points indicate tracking off course or even failed.}
	\label{fig:number_succ}
	\vspace{-10pt}
\end{figure}

\begin{figure*}[t]
	\centering
	\setlength{\abovecaptionskip}{-3pt}
	\includegraphics[width=0.95\linewidth]{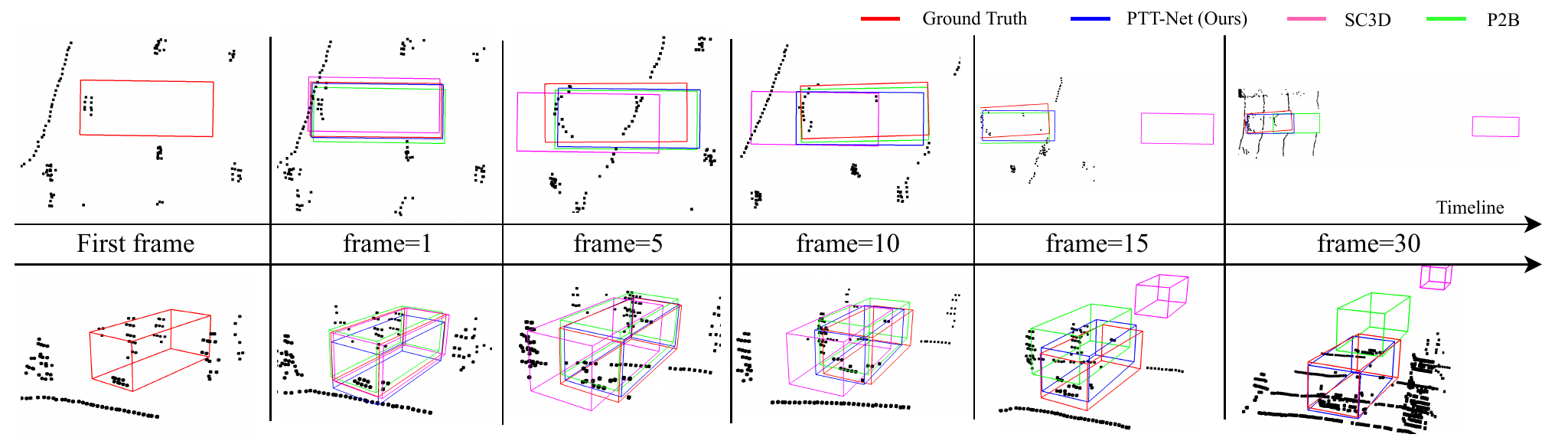}
	\caption{\textbf{Advantageous cases of PTT-Net (ours) compared with SC3D and P2B.} In sparse point cloud scenes, our method can track the target well. Both P2B and SC3D failed to track.}
	\label{fig:qualitative1}
\end{figure*}
\begin{figure*}[t]
	\centering
	\setlength{\abovecaptionskip}{-10pt}
	\includegraphics[width=0.9\linewidth]{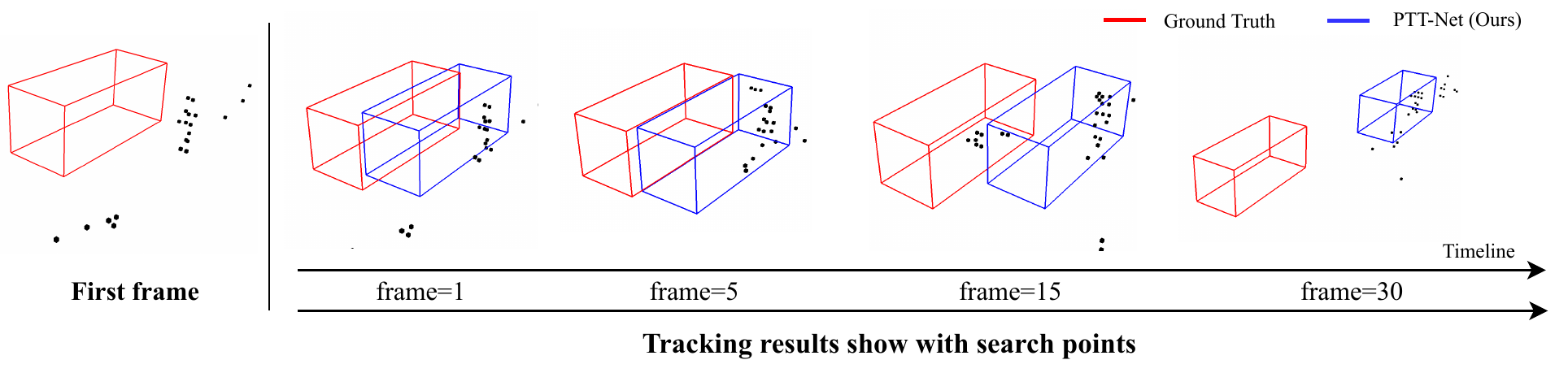}
	\caption{\textbf{Failure cases.} no points in the initial search area}
	\label{fig:failure}
\end{figure*}

\subsection{Qualitative Experiment}
Attention can be understood as the place where the network focuses on. It is obvious that PTT module guides tracker in focusing on foreground points even if they are few.

\subsubsection{Advantageous cases}
We first exemplified the attention score in the voting stage in PTT-Net in Fig.~\ref{fig:attention-show}. To further demonstrate the performance of our method, we selected four scenarios according to the type and difficulty of the tracking target. It is obvious that PTT module guides tracker in focusing on foreground points even if they are few by comparing (a) and (b). Besides, (c) and (d) also show good performance in non-rigid object (pedestrian) tracking, even if the target and background are extremely similar (d), PTT module can distinguish them.

We then visualized our advantageous cases over P2B and SC3D in Fig.~\ref{fig:qualitative1}. We can observe that both SC3D and P2B tracked eventually failed in the sparse scenarios (less than 50 points), but our PTT-Net tracks the target tightly.

\subsubsection{Failure cases}

To show the performance of our method in more detail, we exemplified the impact of varying density of points on the first frame's car to PTT-Net in Fig.~\ref{fig:number_succ}. The orange points indicate failure cases that the number of points from the initial frame are mostly less than 20. And even if four tracklets are initialized with 0 points. In these cases, our PTT-Net could not learn effective object characteristics, so it fails to track. As shown in Fig.~\ref{fig:failure}, our PTT-Net could not learn effective object characteristics since no points in the initial search area.

\subsection{Timing Breakdown}

We calculated the average running time of all test frames in the Car category to evaluate running speed. PTT-Net achieved 40 FPS on a single NVIDIA 1080Ti GPU, including 8.3 ms for preparing point cloud, 16.2 ms for model forward propagation, and 0.5 ms for post-processing. The running time of SC3D \cite{SC3D}, P2B \cite{P2B} and 3D-SiamRPN \cite{3DSiamRPN} on the same platform are 1.8FPS, 45.5FPS and 20.8FPS, respectively.

\section{Conclusions}

In this work, we explored the application of transformer network in 3D SOT task and proposed PTT module. The PTT module aims at weighing point cloud features to focus on the important features of objects. We also embedded the PTT modules into the open-source state-of-the-art method \cite{P2B} and construct a novel 3D SOT tracker named PTT-Net. Experiments show that PTT-Net outperforms previous state-of-the-art methods with remarkable margins. We hope that our work will inspire further investigation of the application of transformers to 3D object tracking.

\section*{ACKNOWLEDGMENT}
This work was supported by National Natural Science Foundation of China (62073066, U20A20197), Science and Technology on Near-Surface Detection Laboratory (6142414200208), the Fundamental Research Funds for the Central Universities (N182608003), Major Special Science and Technology Project of Liaoning Province (No.2019JH1/10100026), and Aeronautical Science Foundation of China (No. 201941050001).


\bibliographystyle{IEEEtran}
\end{document}